\documentclass[letterpaper, 10 pt, conference]{ieeeconf}  

\IEEEoverridecommandlockouts                              

\usepackage{makecell}
\usepackage{multicol}
\usepackage{algorithm2e}
\usepackage{amssymb}
\newtheorem{lemm}{Lemma}
\newtheorem{theorem}{Theorem}
\usepackage{amsmath}
\usepackage{stfloats}
\usepackage{xcolor}
\usepackage{dsfont}
\usepackage{graphicx}
\usepackage{wrapfig}
\usepackage{hyperref}

\title{\LARGE \bf
MSVIPER: Improved Policy Distillation for Reinforcement-Learning-Based Robot Navigation
}

\author{Aaron M.~Roth$^{1,2,}$*, Jing Liang$^{1}$, Ram Sriram$^{2}$, Elham Tabassi$^{2}$, and Dinesh Manocha$^{1}$
\thanks{* Corresponding Author: \url{amroth@umd.edu}}%
\thanks{$^{1}$ Department Computer Science, University of Maryland, College Park, MD}%
\thanks{$^{2}$National Institute for Standards and Technology 100 Bureau Drive, Gaithersburg, MD 20899}%
}

\begin{document}

\maketitle
\thispagestyle{empty}
\pagestyle{empty}

\begin{abstract}

We present Multiple Scenario Verifiable Reinforcement Learning via Policy Extraction (MSVIPER), a new method for policy distillation to decision trees for improved robot navigation.  MSVIPER learns an ``expert'' policy using any Reinforcement Learning (RL) technique involving learning a state-action mapping and then uses imitation learning to learn a decision-tree policy from it. 
We demonstrate that MSVIPER results in efficient decision trees and can accurately mimic the behavior of the expert policy.
Moreover, we present efficient policy distillation and tree-modification techniques that take advantage of the decision tree structure to allow improvements to a policy without retraining.
We use our approach to improve the performance of RL-based robot navigation algorithms for indoor and outdoor scenes. We demonstrate the benefits in terms of reduced freezing and oscillation behaviors (by up to 95\% reduction) for mobile robots navigating among dynamic obstacles and reduced vibrations and oscillation (by up to 17\%) for outdoor robot navigation on complex, uneven terrains.

\end{abstract}

\section{INTRODUCTION}\label{sec:intro} 

Learning methods are increasingly being used for robot navigation in indoor and outdoor scenes. These include deep reinforcement learning (DRL) methods~\cite{chen2017decentralized,drl_navigation,li2019drl_exploration,long2018towards,sathyamoorthy2020densecavoid,liang2021crowd,roth2021XAIN} and learning from demonstration~\cite{argall2009survey,silver2010learning}. These methods have been evaluated in real-world scenarios,  including obstacle avoidance, dynamic scenes, and uneven terrains. 
In these applications, the performance of the navigation methods varies based on the underlying learned policy. It turns out that the learned policy may have errors or be otherwise sub-optimal. Additionally, the learned policy is often represented in the form of a neural net, which can be rather difficult to analyze.

In this paper, we are mainly interested in developing approaches to Reinforcement Learning that use a decision-tree (DT) policy.
A decision tree can be more easily analyzed and modified, and the cause-and-effect of state and action is apparent.
Learning a successful tree policy is typically more difficult than learning with a neural net. Moreover, these issues are compounded when one attempts to produce a policy
for use beyond toy problems and proof-of-concepts~\cite{frosst2017distilling,curram1994neural,Roth_MS_Thesis2019}.
We further investigate using the tree policy to enable modifying and improving upon a policy when the original learning pipeline becomes unavailable.  In the real world, there will be situations where we want to take a learned policy for an autonomous robot, and analyze its shortcomings (\textit{including} shortcomings defined or noticed after initial training!) and then fix those shortcomings. Retraining will not always be possible, nor will the original dataset always be available.

We introduce a novel method, Multiple Scenario Verifiable Reinforcement Learning via Policy Extraction (MSVIPER), and use it for improved robot navigation in complex indoor and outdoor scenes. Our formulation extends VIPER~\cite{VIPER} and learns an ``expert'' policy using a neural net (such as PPO~\cite{PPO}) before using imitation learning to fit a DT to replicate the expert policy.  
MSVIPER compensates for some of VIPER's limitations and also adds new capabilities. There has been limited work on applying VIPER to more complex tasks and almost no work on applying it to any real-world applications. Even for the tasks to which it has been applied, there is no recourse for addressing any errors or issues discovered.

 \begin{figure}[t]
    \centering
    {{\includegraphics[width=0.1955\textwidth,height=0.1445\textwidth]{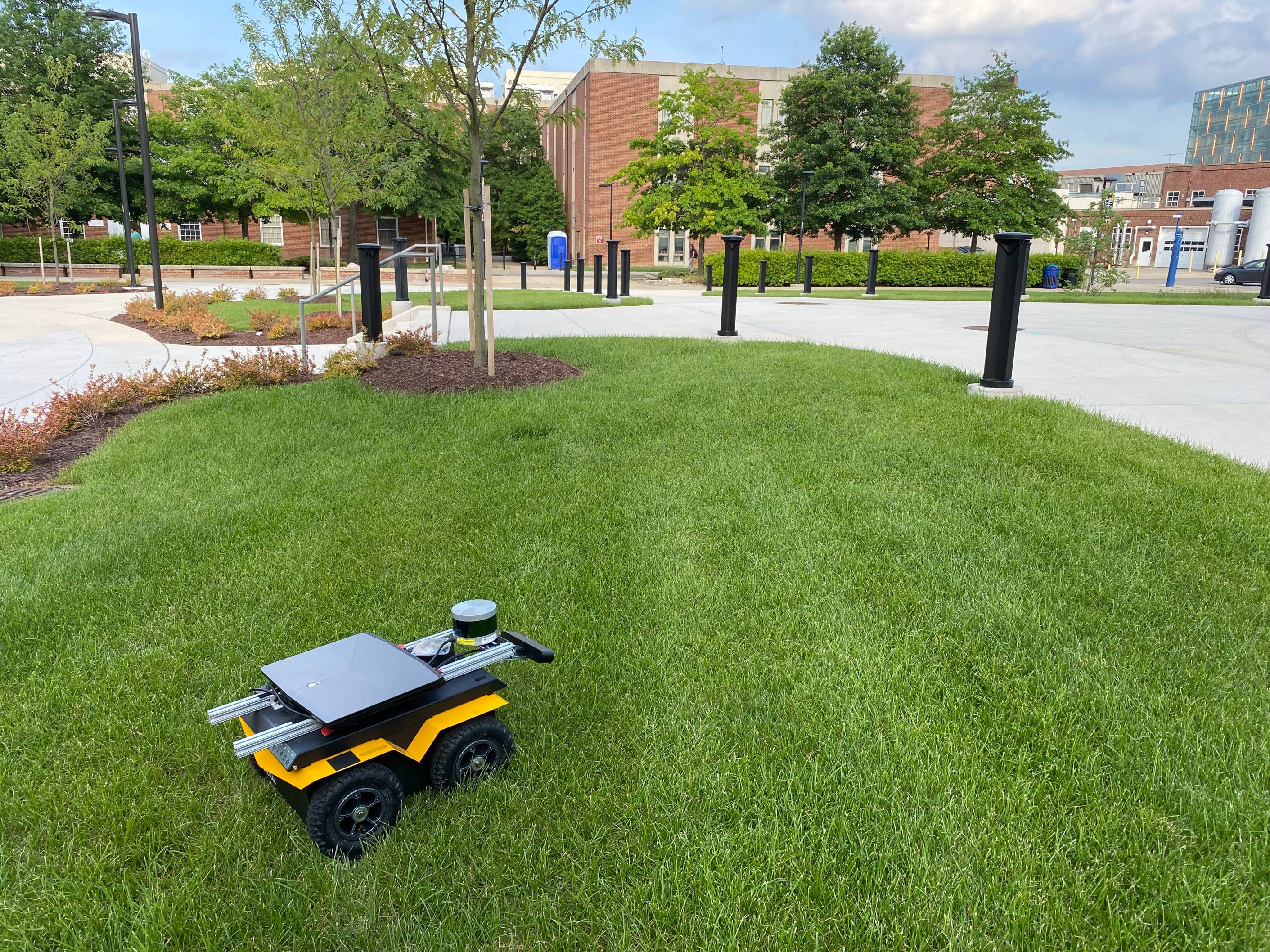} }}%
    {{\includegraphics[width=0.1955\textwidth,height=0.1445\textwidth]{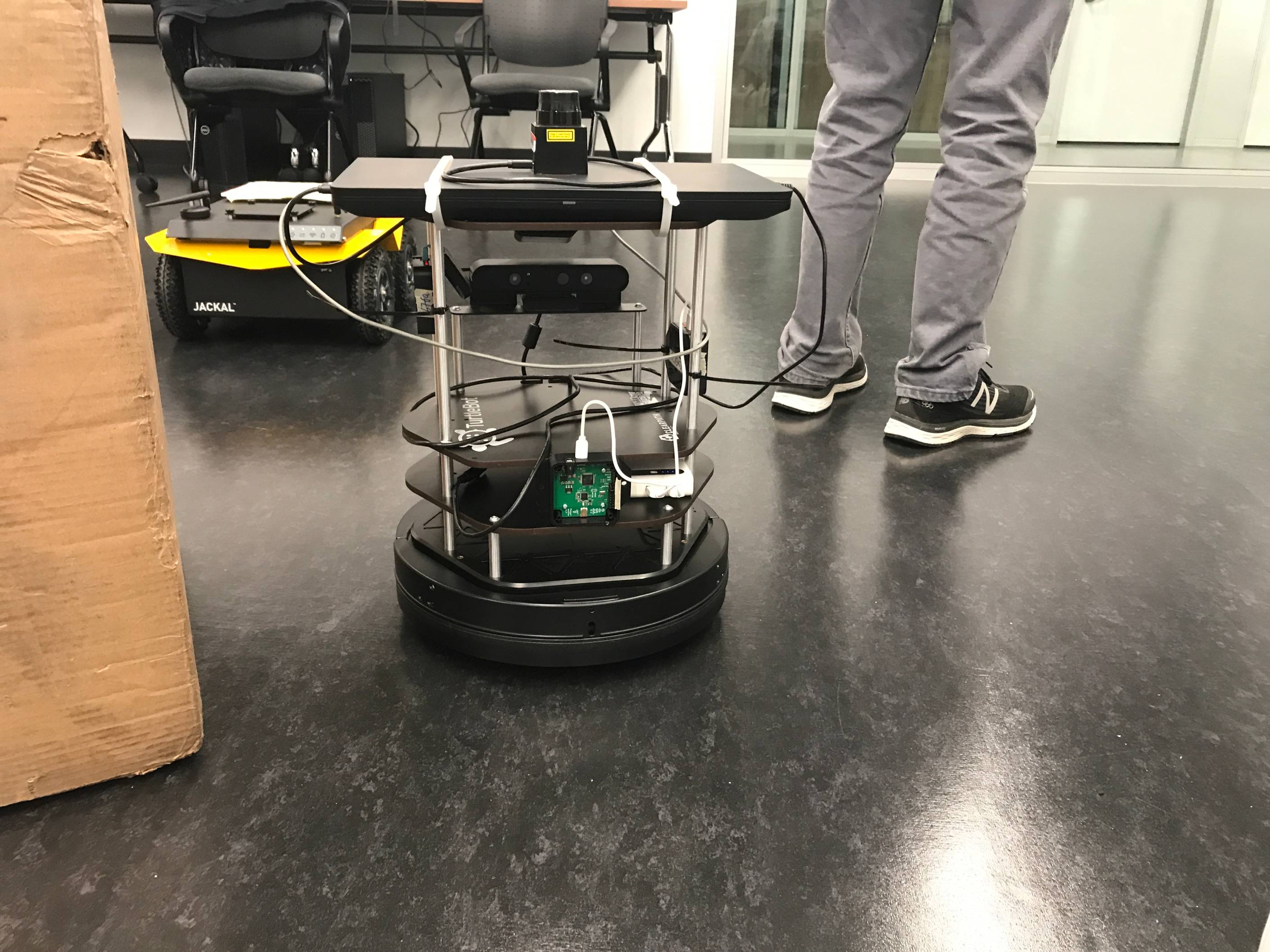} }} 

    \caption{
    MSVIPER: The left image is from the uneven outdoor terrain environment navigation. Our policy distillation and tree modification reduce oscillation and vibrations in the resulting trajectory;
    the right image is a test of the real-world tree policy for indoor obstacle avoidance, where MSVIPER results in smoother trajectories and overcomes freezing problems. 
    }
    \label{fig:frontpage}
\end{figure}

We use MSVIPER to improve the performance of reinforcement-learning-based robot navigation methods. Instead of using different heuristics or trying different parameters and repeating the time-consuming training step, we integrate expert policies, policy extraction, and decision trees. Our {\em policy distillation} approach directly analyzes  and performs tree modification operations on the DTs. 
Some of the novel contributions of our work include:

\begin{enumerate}
    \item We present MSVIPER to improve imitation learning of DTs for indoor and outdoor robot navigation policies. MSVIPER samples from trajectories in multiple environmental scenarios. 
    MSVIPER also results in smaller DTs, which are  easier to understand and analyze~\cite{quinlan1987simplifying}. We also demonstrate the superior sampling complexity of MSVIPER. 
    \item We describe techniques to modify and improve an initial tree policy \textit{without retraining}, which could not be performed using a standard neural net policy. 
    We use these improved tree policies to generate smooth paths for collision-free robot navigation, to fix freezing problems in crowded indoor scenarios, and to reduce vibrations and oscillations on complex outdoor terrains.
    \item We highlight the performance of improved policies on 
    indoor and outdoor navigation scenarios. The indoor navigation is performed using a Jackal robot and Turtlebot among moving pedestrians and we observe up to $95\%$ improvement in terms of reduction in freezing and oscillation behaviors. The outdoor navigation is performed using a Clearpath Husky and Jackal on uneven terrains with obstacles. We observe up to $17\%$ improvement in terms of reduced vibrations and oscillations.  
\end{enumerate}

\section{BACKGROUND AND RELATED WORK}\label{sec:rw} 

There is considerable work on using machine learning methods for robot navigation, interpretation of networks~\cite{doshi2017towards} and decision trees~\cite{Roth_MS_Thesis2019}. 
A soft decision tree can be distilled from a deep neural network~\cite{frosst2017distilling} or learned directly via a Policy Tree~\cite{das2015adaptive}.
A tree can be learned in an additive manner using a reinforcement-learning-style approach~\cite{pyeatt2003reinforcement} and Conservative Q-Improvement ~\cite{roth2019conservative}. 

Policy distillation involves transformation  of a policy from one format to another while keeping the essential input-output as similar as possible. This could be transforming a neural net or Markov Decision Process (MDP) into a smaller network or MDP~\cite{delgrange2022distillation}, or into a different form entirely, such as a saliency map~\cite{xing2022policy} or tree~\cite{li2021neural,VIPER}.
For policy distillation, our approach is based on
VIPER~\cite{VIPER}  because it is flexible in terms of being able to use any method to learn the expert policy. 
In practice, VIPER's usage has been limited to proof-of-concept problems such as Cartpole~\cite{bhupatiraju2018towards}, 
Pong~\cite{VIPER} and other simulations such as CARLA~\cite{chen2020learning}.

Autonomous systems need the ability to deal with uncertain and unforeseen scenarios.
Some existing systems reduce risk or correct errors by gathering information from a human through methods such as learning from demonstration~\cite{brown2018risk,cai2021vision,brown2017toward}.
Human feedback can also incorporate missing state features into state representation~\cite{basich2020improving}. 
One approach asks a user to label states of interest as a means of gaining information during training~\cite{singh2019end}. 
The above-mentioned approaches incorporate human expertise while remaining black boxes. They typically require retraining, which can be time consuming. 
In our approach, we can evaluate the effectiveness and efficiency of the learned policy and directly improve without retraining.

\section{PROBLEM FORUMUATION AND NOTATION}\label{sec:notation}
As is typical in approaches that make use of reinforcement learning methods, we model the underlying task in the application as a Markov Decision Process (MDP), which could be represented by a 5-tuple $(S, A, P, R, \gamma)$. $S$ is state space, $A$ is action space, $R$ is rewards, and $P$ is state transition dynamics: $S \times A \xrightarrow{} S$, and $\gamma \in (0,1 ] $ is the discount factor.  Distribution of initial state is represented as $\rho_0$. We want to generate an optimal policy $\pi$ that maximizes discounted reward function:
\begin{equation*}
        \eta(\pi) = \mathbb{E}_{\pi}\left[\sum^{T-1}_{t=0}{\gamma^{t} r(s_t,a_t)}\right]. 
\end{equation*}
where $T$ is the time horizon. 
The policy is represented by networks with parameter $\theta_\pi$. 

\begin{figure*}
    \centering
    \includegraphics[width=0.8\textwidth]{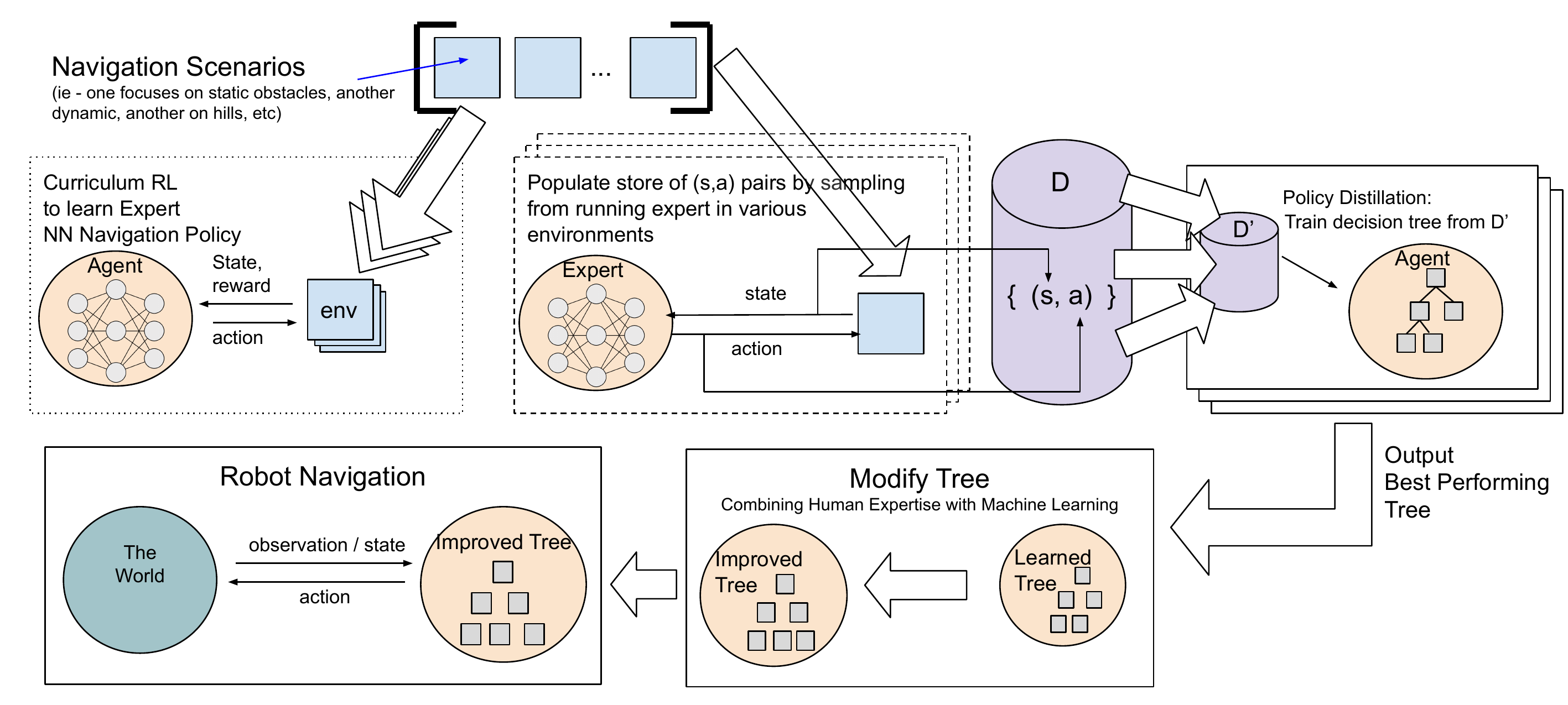}
    \caption{Overview of our Policy Distillation and Tree Modification: We highlight three parts of MSVIPER in the top row. First, we train an expert policy with different scenarios. Second, we generate trajectories using the expert policy.
    Third, we use the trajectories to train a decision tree. 
    Next, we use our tree modification techniques to correct errors and improve navigation performance without retraining (bottom row).
    }
    \label{fig:pipeline}
\end{figure*}

In the Deep RL learning step, we want the robot to learn to accomplish one of the three given tasks, involving navigating to a static or dynimc goal position, avoiding collisions with static or dynamic obstacles, and achieving a smooth trajectory that avoids vibrations and oscillations.  

This policy is black-box, it may be suboptimal, and it may have errors a certain percentage of the time. We seek to address such issues during the distillation and modification procedures described in Section \ref{sec:app}. 

We assume the researcher or developer working with the model has access to the model itself in the form of a black box, but does not have the capability to retrain it (simulating technological, temporal, organizational, legal, or other constraints that arise in the real world).

The notation we use is shown in Table \ref{tbl:notation}.
\begin{table}[h!]
    \centering
    \resizebox{0.45\textwidth}{!}{
    \begin{tabular}{|c|l|}
\hline
$s \in S$ & \makecell[l]{a state $s$ is an array representing the state \\ of the world and the robot in it} \\
\hline
$a \in A$ & \makecell[l]{an action $a$ is a single discrete action in the \\ set of possible actions $A$} \\
\hline
$P$ & state transition dynamics\\
\hline
$r, R$ & reward (for a single ($s$, $a$) pair or in general, respectively)\\
\hline
$\gamma$ & future discount factor\\
\hline
$\pi$ & a policy (whether neural net or tree based)\\
\hline
$\pi^*$ & an expert policy (neural net)\\
\hline
$\hat{\pi}$ & a decision tree policy\\
\hline
$l_t$ & length of trajectory to use when generating state-action pairs\\
\hline
$M$ & number of trajectories used while generating state-action pairs\\
\hline
$N$ & number of VIPER iterations used to run per scenario\\
\hline
$E$ & \makecell[l]{ an ordered list of environments/scenarios \\ (repetition is allowed)}\\ 
\hline
$n_{cv}$ & \makecell[l]{ the number of cross-validation trials used to \\ compare the candidate final policies $\hat{\pi}$}\\
\hline
$D, D_i, D'$ & unordered sets of state action pairs ( $\{(s, a)\}$)\\
\hline
$n_s$ & number of samples to use when generating $D'$ from $D$\\
\hline
$e_O$ & \makecell[l]{ Objective Efficiency, a measure of how efficiently a tree \\ modification improved a policy; is higher when the target metric\\ has higher improvement per node-modified} \\
\hline
$e_R$ & \makecell[l]{  Relative Efficiency, a measure of how much a tree\\  modification improved a policy; is higher when the target metric \\
has higher improvement per percentage-of-nodes-modified} \\
\hline
$M_1, M_2$ & values of a target metric before and after tree modification\\
\hline
$N_+$ & \makecell[l]{Number of nodes added, removed, or modified in a tree\\ modification procedure} \\
\hline
$N_1$ & Number of nodes in $\hat{\pi}$ before modification \\
\hline
\end{tabular}
}
    \caption{Symbols and notation used in the paper}
    \label{tbl:notation}
\end{table}

\section{OUR APPROACH: MSVIPER}\label{sec:app}
We present a pipeline in Figure \ref{fig:pipeline}, the end-result of which is a modifiable decision tree policy for robot navigation. 
Section \ref{sec:app-rl} discusses use of deep reinforcement learning to create an initial learned navigation policy. Section \ref{sec:app-dt} describes policy distillation into a decision tree using MSVIPER. Section \ref{sec:app-mod} describes our tree modification procedure to improve the policy for robot navigation. We highlight its benefits in terms of addressing the issues related to freezing, oscillation, and vibrations in a robot's trajectory. 

\subsection{Reinforcement Learning for Initial Navigation Policy}\label{sec:app-rl} 

We solve the Deep RL task as formulated in In Section \ref{sec:notation}.
We trained using a standard Deep RL algorithm such as the Proximal Policy Optimization (PPO)~\cite{PPO} algorithm.
One important detail to note is that we trained our agent in a succession of increasingly difficult environment stages and not in a single stage. 
Many methods use Curriculum Learning (CL)~\cite{bengio2009curriculum} to learn good navigation behavior. In CL, instead of a single environment, the agent is trained in a succession of increasingly difficult environment stages. In CL, the policy trained by each scenario is reused as an initialization of the next scenario of training.  In particular, the final policy from an earlier scenario becomes the starting policy for a later scenario.  We used CL in the Deep RL stage,
creating multiple scenarios of increasing difficulty in each case and use them to train a neural net format policy.

\subsection{Policy Distillation: Conversion to Decision Tree Policy}\label{sec:app-dt} 

We perform policy distillation on the initial policy to create a DT policy. 
First, we generate trajectories, which are sequences of state-action pairs ($\{(s_0, a_0), ...,  (s_i, a_i)\}, s \in S, a \in A$) in a manner that is more sample efficient for some environments than previous approaches.
Prior methods (VIPER~\cite{VIPER}) would sample trajectories generated by applying the policy in a single simulated environment, and use those trajectories to train a decision tree. We refer to it as single source VIPER (SSVIPER).
Instead, we sample the trajectories generated by applying the policy in multiple simulated environments of varying complexities or types. 
We try to ensure a large variety in terms of generating indoor and outdoor simulation environments. This variety
 ensures that the DTs that result from learning based on the sampled pairs will be more robust to different scenarios that might be encountered in the real world scenarios.


\RestyleAlgo{ruled}
\LinesNumbered
\begin{algorithm}
\textbf{MSVIPER}( $(S, A, P, R), \pi^*, M, N, E, l_t, n_{cv}, n_s$):\\
$D \gets \varnothing$ \;
\For{ scenario $e \in E$}{
    \For{ $i = 1 \text{ to } N$}{
        $s_0 \gets$ initial state after resetting $e$\; 
        $D_i \gets \{ (s_j, \pi^*(s_j)) \sim d^{s_{j-1}} \}  $ (sample $M$ trajectories of length $l_t$, by running policy $\pi^*$ in environment scenario $e$)\;
        $D \gets D \cup D_i $\;
        $D' \gets \{(s, a) ~ \{(s, a)\} \in D\} $ (sample a dataset, retrieve a sample size of $n_s$ pairs)\;
        Train decision tree $\hat{\pi}_i$ using $D'$ as dataset\;
    }
}
Return best policy $\hat{\pi}_i \in \{\hat{\pi}_1 \text{ to } \hat{\pi}_N \}$ using $n_{cv}$ trials to check each of them.\\
 \caption{Multi-Scenario VIPER: \newline 
 Our improved algorithm for learning a decision tree policy through imitation learning of an expert policy across multiple scenarios
 \label{alg:msviper}}
\end{algorithm}

In MSVIPER (Algorithm \ref{alg:msviper}), 
for each scenario $e \in E$, we use this procedure:
For a sequence of $N$ iterations, we sample trajectories of length $M$. 
All such pairs are put into a set. Pairs are randomly drawn from that set and used as a dataset to train a DT in a supervised learning manner using CART~\cite{lewis2000introduction}.
The overall dataset from which pairs are drawn is maintained between the stages. 
The resulting $N$ decision trees are evaluated in simulation for $n_{cv}$ trials, and the best-performing (highest reward) tree is the output of the method. 


\subsection{Tree Modification: Policy Optimization \& no Retraining}\label{sec:app-mod} 

\nocite{roth2021XAIN}

We use our tree-based policy for analysis, verification, and modification.
We modify the tree to improve the navigation characteristics of the policy without retraining.

A key issue in terms of tree modification is to change as few nodes as possible. 
Details of our novel algorithms for different types of modifications to improve navigation can be found in 
Appendix \ref{sec:app-mod}
, and are summarized below.
Tree modifications can include changing the action on a leaf node, changing the feature a node splits on or the threshold at which it splits, removing nodes, or adding new nodes.

\subsubsection{Indoor Navigation: Tree Modification}\label{sec:app-mod-indoor} 

We address two main issues with respect to learning-based methods used for indoor navigation: freezing robot problems and oscillations.
A common issue in terms of robot navigation in indoor scenes among dynamic obstacles is the ``Freezing Robot Problem''~\cite{trautman2010unfreezing,trautman2015robot,sathyamoorthy2020frozone,fan2019getting}. It corresponds to configurations, where
the robot is presented with a set of obstacles and chooses to remain immobile to avoid any collision.
We modify the policy to reduce such instances.
Our freezing identification procedure identifies nodes in a decision tree that cause (or could potentially cause) the freezing behavior and change the attributes of the identified nodes.
The algorithm analyzes each leaf node in the tree. If the node's action is the Stop action, and if the occupancy grid described by the state bounds of that node indicate stationary obstacles (within a configurable tolerance for minimal movement), then the node is considered to be a problematic node and is added to the list of potential-freezing nodes. Each potential-freezing node is adjusted, given a rotation action.

We also modify the tree to generate smoother trajectories and reduce the oscillations. The oscillatory behavior can occur when a robot moves around the obstacles. Oscillations can be defined in two ways:
first as a percentage as in $C_{osc} = (c(a_i) / T) \space \forall \space i = \{0, ... T\}$,  where $T$ is a number of timesteps and $c(a_i) = 1$ if the action $a_i$ is part of a sequence of actions of certain length 
that alternate between left-right rotation angular velocity, $c(a_i) = 0$ otherwise; and  
secondly, as $C_\text{osc}=\frac{1}{N}\sum_{i=0}^N abs(\omega_t- \omega_{t-1})$, where $\omega$ is the angular velocity of the robot and the equation indicates the average changes in the angular velocities. We use the simulations to record such behaviors and use that to identify the corresponding potentially oscillating nodes in the tree. We modify the tree by modifying existing nodes or adding nodes {  (depending on calculations of how much of the problem node's subspace is or is not affected by the same problem), replacing existing actions with replacement actions that have lower linear and angular velocities.

\subsubsection{Outdoor Navigation: Tree  Modification}\label{sec:app-mod-outdoor} 

In outdoor navigation, a robot needs to deal with complex, uneven terrains with obstacles. This tends to cause jerky motion with bounces and high vibrations.
%
We define robot vibration as:
\begin{align}
    V_b &= \sum_{t'=t-3}^t \gamma^{t-t'} ( |{\omega_r}|+| \omega_p|) = \sum_{t'=t-3}^t \gamma^{t-t'} w_{t'}
\end{align}
where $\omega = |\omega_r|+|\omega_p|$. $V_b$ is the vibration of the robot, where $\gamma$ is the discount factor and $\omega_r$ and $\omega_p$ are the angular velocities in the roll and pitch directions.
We 
analyze the robot's angular velocity and identify the corresponding potentially vibrating nodes. We look at nodes that split on angular velocity as well as nodes that exist inside or intersect the ``vibration space'' (the closed hyperdimensional object described by $V_b$; this is a hyperrectangle and convex hull). 
Our formulation is able to treat the vibration threshold as a hyperparameter and tune it to decrease the vibration in the learned policy. 
We can modify the tree by adjusting the threshold of certain nodes or by changing the action (and thus motion primitive) to reduce linear and/or angular velocity for that node. 
We are also able to modify the tree to reduce the oscillations, similarly to indoor navigation.

\subsection{Analysis: Upper Bound Divergence}
Compared to SSVIPER, MSVIPER makes smaller and more efficient DTs.

\begin{theorem}
For MSVIPER, the loss function is :
\begin{equation}
        \Tilde{l}_{t}(s,\pi) = V_{t}^{(\pi ^*)}(s)-\min_{a \in A}\hat Q_{t}^{(\pi ^*)}(s,a),
\end{equation}
where
\begin{equation}\label{th:final1}
    \hat Q^{\pi^*}(s,a) = 
    \mathbb{E}_{\pi}
    \left[\sum_{e \in E}\left(\sum_{k=0}^{\infty}\gamma^{k}R^e_{t+k+1}|S_{t}=s, A_{t}=a\right) \cdot w_e\right]
\end{equation}
\label{theorem:loss_function}
\end{theorem}

$\hat Q^{\pi^*}(s,a)$ is the quality function, $V_{t}^{(\pi ^*)}(s)$ is the value function of the policy, $R_i^e$ is reward at timestep $i$ in environment $e$, and $w_e$ weights the contribution of each environment. In order to guarantee improved performance, we need to decrease the upper bound on $\Tilde{l}_{t}(s,\pi)$.  

\begin{lemm}\label{lemma:l1}
    for a given $\pi^{*}$, sampling multiple
    scenarios: 
    \begin{equation}\label{eq:loss_boundance0}
    J(\hat{\pi})\le J(\pi^{*})+T\epsilon_{N}+\Tilde{O}(1)
    \end{equation}
\end{lemm}

where $J(\hat{\pi})=-V(\hat{\pi})(\mathbf{s})$ be the cost-to-goal from the state $\mathbf{s}$. The $\epsilon_N$ is the training loss defined in~\cite{VIPER}. T and N are time steps and total training iteration numbers, respectively. $\Tilde{O}(1)$ is a constant function.  Equation \eqref{eq:loss_boundance0} holds true.

The proof of the theorem is 
Appendix \ref{sec:app-full-proofs}.

\section{Results}\label{sec:results}

\subsection{Environments} 

We tested the performance of MSVIPER in three environments: i) an indoor environment with obstacles for collision-free navigation; ii) an indoor warehouse environment where a robot must find and follow a human, and iii) an outdoor   environment with uneven terrain a robot must traverse. 

In the indoor Obstacle Avoidance environment, the robot starts in a random location and must navigate around static and dynamic obstacles to a random goal location.  
We desire that the policy used for interpretation should perform well in terms of avoiding the pedestrians and obstacles.

In the indoor Find and Follow environment, the robot spawns in a complex multi-room environment with obstacles, a warehouse full of haphazardly arranged boxes. It spawns in a room. It must exit the room and find and follow a human in a white shirt around a warehouse. There are three stages in this task: i) learning the exit the room, ii) learning to exit the room and finding the human, iii) following the human as they move.

The outdoor navigation environment has a starting location and goal location and uneven terrain between them. The robot must navigate over the terrain safely and efficiently. 
There are no obstacles in this environment but the challenge is optimally navigate over inclined or uneven outdoor terrain.

\subsection{MSVIPER: Improved Navigation}\label{sec:results-emp} 

    \begin{table}[t]
    \centering
    {\resizebox{.5\textwidth}{!}{
    \begin{tabular}{|c|c|c|c|c|c|c|c|c|c|}
        \hline
        Environment & \makecell{Policy\\ Type} & \makecell{Final \\ Avg \\ Reward/ts \\ (test)} & \makecell{Size \\ (\# Nodes)} & Depth & Runtime(s)\\
        \hline
        Obstacle Avoidance & Expert NN & 0.29 & n/a & n/a & 7.81\\
        Obstacle Avoidance & SS VIPER & 0.23 & 2651 & 25 & 8.84 \\
        Obstacle Avoidance & MSVIPER & \textbf{0.30} & 411 & 16 & 7.59\\
        \hline
        Find \& Follow & Expert NN & 0.464 & n/a & n/a & 90.3726\\
        Find \& Follow & SS VIPER & 0.097 & 2073 & 22 & 51.00\\
        Find \& Follow & MSVIPER & \textbf{0.19}  & \textbf{1681} & 20 & 34.02\\
        \hline
    \end{tabular}
    }}
    \caption{ 
    Improved Navigation using MSVIPER
    }
    \label{tab:msviper}
\end{table}

\begin{table*}[t]
    \centering
    \resizebox{1.0\textwidth}{!}{
    \begin{tabular}{|c|c|c|c|c|ccc|c|c|}
        \hline
        \makecell{Indoor Navigation Tree Policy} & $n_s$ & $N_1$ & $N_2$ & $N_+$ & \makecell{Freezing \%:} &  (initial) & (modified) & $e_O$ & $e_R$ \\
        \hline
        Policy with Freezing Error & 5000 & 319 & 319 & 12 && 100 & 5 & 0.079 & 25.25 \\
        \hline
        & & & & & \makecell{Oscillation \%} & (initial) & (modified) & & \\
        \hline
        Policy with Excessive Oscillation & 30,000 & 1429 & 1429 & 11 && 96 & 6 & 0.085 & 121.7 \\
        \hline
    \end{tabular}
}
    \caption{Tree Modification for Indoor Robot Navigation}
    \label{tbl:tree_mod_indoor}
\end{table*}

\begin{table*}[t]
    \centering
    \resizebox{1.0\textwidth}{!}{
    \begin{tabular}{|c|c|c|c|c|ccc|c|c|}
        \hline
        \makecell{Outdoor Navigation Tree Policy} & $n_s$ & $N_1$ & $N_2$ & $N_+$ & \makecell{Vibration: } &  (initial) & (modified) & $e_O$ & $e_R$ \\
        \hline
        Policy with Excessive Vibration & 10,000 & 17 & 17 & 1 & & 0.53 & 0.48 & -0.084 & 8.29 \\
        \hline
        & & & & & \makecell{Oscillation \%} & (initial) & (modified) && \\
        \hline
        Policy with Excessive Oscillation & 10,000 & 17 & 17 & 2 & & 0.092 & 0.076 & 0.089 & 1.51 \\
        \hline
    \end{tabular}
    }
    \caption{
    Tree Modification for Outdoor Robot Navigation
    }
    \label{tbl:tree_mod_outdoor}
\end{table*}

MSVIPER achieves a higher average reward, smaller trees, and faster runtime than SS VIPER. 
%
%
As shown in Table \ref{tab:msviper} we compared the 
performances by expert policies in simulation using MSVIPER and SSVIPER.
MSVIPER generated DTs for the indoor navigation obstacle avoidance environment had a simulated reward near the expert policy level. Moreover, it has less nodes (smaller size) than the SSVIPER policy.
MSVIPER exhibits better performance than SSVIPER. In the obstacle avoidance environment, 
The robot trained by MSVIPER has a better chance of going to the goal without any collisions. 
MSVIPER results in a smaller tree than VIPER, making it tree modification easier.
In the Find and Follow scenario,  MSVIPER results in smaller DT and improves the final average reward compared to SSVIPER. The superior reward for MSVIPER also indicates shorter navigation time for the robot.
In all indoor environments, MSVIPER takes advantage of the curriculum stages to increase sample diversity and cover more critical states. It also reduces the chances of overfitting.

Figure~\ref{fig:sample_complexity_size} shows the sizes of trees as sample complexity changes. More samples result creation of trees that more closely mimic the expert policy. MSVIPER results in a smaller size tree than VIPER and that makes it easier in terms of tree modification.

\subsection{Indoor Navigation Results} 

For indoor navigation, we validated our learned decision tree policy's ability to handle navigation in a real-world environment using two robot platforms, Clearpath Jackal~\cite{clearpath_jackal} and Turtlebot.  We tested the policy on both robots in lab environments with various furniture and moving humans serving as obstacles. We observe significant reductions in the freezing behaviors with dynamic obstacles and fewer oscillations using MSVIPER and tree modifications (see Table \ref{tbl:tree_mod_indoor}.  We observe significant improvements, by $90-95\%$, by comparing the paths computed by the initial navigation policy verses the modified policy.

\subsection{Outdoor Navigation Results}\label{sec:results-outdoor} 

We tested our trees and also the expert policy in the real-world outdoor environment with uneven terrain using a Clearath Husky robot. The trained trees have very similar performances to the expert policy. After the tree modification the vibration and oscillation of the robot both reduce and the robot runs much more stably and smoothly.  We highlight the benefits of policy distillation and tree modification in terms of reduced vibrations and oscillations on uneven outdoor terrains by $17\%$ (see Table \ref{tbl:tree_mod_outdoor}).

\begin{figure}[t]
    \centering 
    \includegraphics[width=0.4\textwidth]{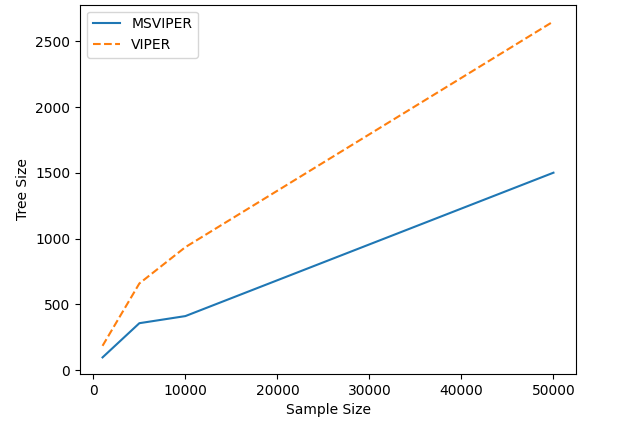}
    \caption{Size of Trees Generated for VIPER and MSVIPER.  MSVIPER produces smaller trees for the same total number of samples, on average.
    Smaller trees are beneficial from an interpretability standpoint~\cite{quinlan1987simplifying}.}
    \label{fig:sample_complexity_size}
\end{figure}

\begin{figure}[b]
    \centering 
\includegraphics[width=0.2\textwidth,height=0.17\textwidth]{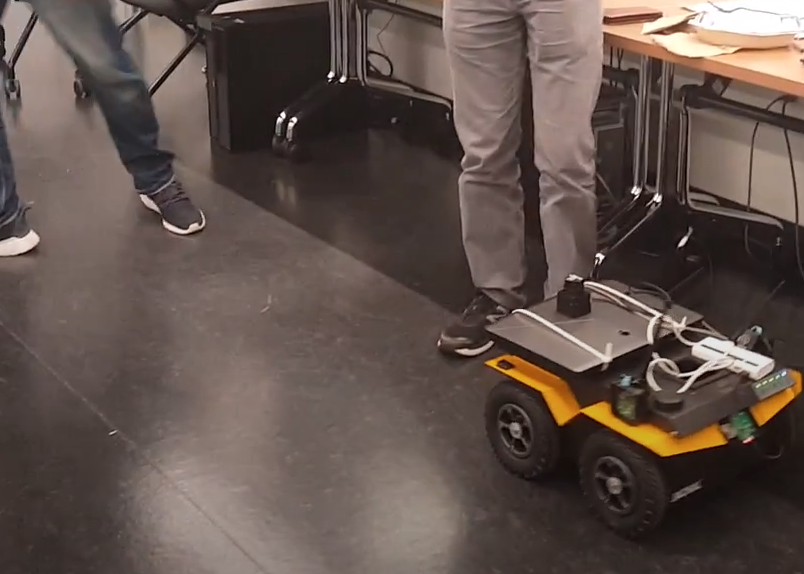}
    \includegraphics[width=0.2\textwidth,height=0.17\textwidth]{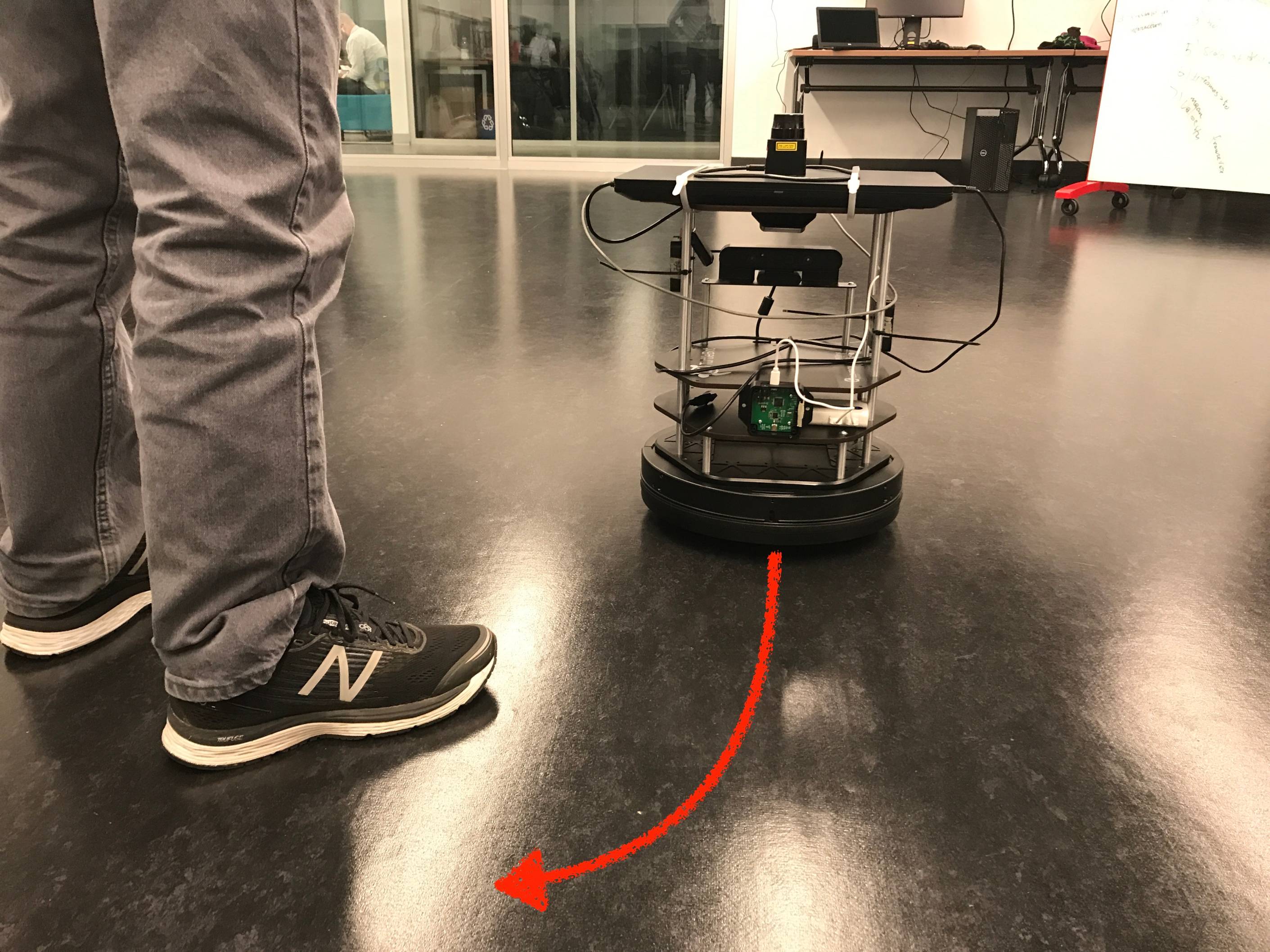}
    \caption{We have evaluated MSVIPER on a Jackal Robot and a Turtlebot navigating in real-world scenarios with pedestrians and obstacles.  The 
    arrow indicates the path taken by the robot as a result of the policy, showing it successfully navigating around the obstacles. We observe fewer freezing and oscillatory behaviors using MSVIPER.
    }
    \label{fig:realworld}
\end{figure}

\subsection{No-Retraining Tree Modification: Efficiency Metrics}\label{sec:results-mod} 

First, for each navigation behavior (freezing, oscillation, vibration) we create a domain-specific procedure that can detect whether a given tree policy contains nodes that could results in those behaviors, and which detects those nodes. We are able to perform this analysis with a DT, but would be much harder with  a neural net structure. 
Second, given the above,
we can modify the policy to mitigate or fix the issue. 
We evaluate the policies according to navigation metrics before and after the trees are modified, to determine whether there was improvement. We can also measure the improvement with tree modification efficiency metrics.

In particular, we evaluate the benefits of tree modification using two metrics.
Table \ref{tbl:tree_mod_indoor} and \ref{tbl:tree_mod_outdoor} show the \textbf{Objective Efficiency} $e_O$ and the \textbf{Relative Efficiency} $e_R$ of the modifications. 
Each metric is formulated with respect to a target metric $M$ that the tree modification is attempting to accomplish. In our demonstrations, $M$ is the percentage of trials where freezing occurs (for freezing), 
$C_{osc}$
(for oscillation), and 
$V_b$ (for vibration).
If we need to make a larger change in the initial policy to achieve the desired change in the target metric, it results in lower efficiency. These metrics are computed as:
\begin{equation}
    e_O = \frac{|M_2 - M_1|/M_1}{N_+}, \quad e_R = \frac{|M_2 - M_1|/M_1}{N_+/N_1},
\end{equation}
where $M_1, M_2$ are the values of the target metric (either \%-of-trials-where-freezing-occurs or \%-of-trials-where-oscillation-occurs) before and after tree modification, respectively, $N_1$ is the number of nodes before modification, and $N_+$ is the number of nodes modified (changed or added).  
As shown, $e_O$ is higher when the target metric has higher improvement per node-modified, and $e_R$ is higher when the target metric has higher improvement per percentage-of-nodes-modified.

For example, in the best-case freezing fix for MSVIPER, running the freezing-detection algorithm revealed that 30 out of 199 nodes of the tree were configured such that they could lead to a freezing issue in the initial learned policy. 

We analyzed many aspects of computing efficient DTs and tree modifications.
One additional benefit of the tree modification metric, in addition to quantifying improvement, is higher efficiency of MSVIPER over SSVIPER. For oscillation reduction in  indoor navigation, the tree improved using our MSVIPER method has a higher objective and relative efficiency than that improved using SSVIPER (the MSVIPER $e_O$ and $e_R$ are shown in Table \ref{tbl:tree_mod_indoor}, VIPER has $e_0 = 0.032, e_R = 32.9$, both lower that MSVIPER's). Since the initial MSVIPER policy closely mimics the expert policy, only 11 nodes require modification to fix the oscillation, as compared to 31. To improve the freezing behavior, 
the most-improved SSVIPER policy had an $n_s$ of size 5K (shown in the Table) compared to the best performing MSVIPER at $n_s=$ of size 1K. Freezing behavior was not present in \textit{all} initial policies. MSVIPER at a 1K sample size resulted in no freezing behavior.   In other words, MSVIPER resulted in a policy with fewer errors from fewer samples, and there were fewer errors to correct in the best-case policy (in the case of freezing). This highlight many benefits of MSVIPER in terms of efficient DTs and policy distillation.

\subsection{Discussion of Implications}

We have shown that policies from MSVIPER can be improved by tree modification without retraining.  
This is of critical importance in real-life applications.  
In some applications, we are often interested in the initial training process.  In scenarios where the users will train and use RL models, however, there is great utility in a pipeline that allows for modifying a policy after-the-fact. 
This has applicability in situations where training is prohibitively expensive, for example, as well as situations where customization is useful. In particular, it allows improving and modifying policies by a software engineer or other developer without specific knowledge of reinforcement learning.

The examples involved here involved improving upon a particular metric, correcting an error, and even converting a failing policy into a successful one. One can imagine application developers creating solutions to error-prone or optimal learned policies and improving upon them further, 
delivering results beyond those that could be achieved by man or machine alone. 

\section{CONCLUSION, LIMITATIONS, FUTURE WORK}\label{sec:conclusion} 

We introduce MSVIPER, which learns a decision-tree-style policy for complex navigation tasks by imitating a neural-net style policy and enable post-training modifcation. We demonstrate the validity of approach in  simulation environments and in real-world scenarios in the context of a robot navigating obstacles in a crowd, a robot following a person in a warehouse and robot navigation in complex outdoor terrains.
%
We show examples of our policy structure allows it to be modified and improved without retraining.

Limitations exist: MSVIPER requires a discrete action space, a numerical-array-format state space, and hyperparameter tuning (its predecessors share these limitations).
Our approach is best suited for scenarios where training can be expensive and it would be useful to reduce the overhead of retraining with different parameters. 

Future work could involve using MSVIPER for applications of reinforcement learning beyond robot navigation. Additionally, our tree modification could be used on trees generated via other means.




\bibliographystyle{IEEEtran.bst}
\bibliography{biblio.bib}  



\newpage
\newpage
{\color{white}.}
\newpage
\appendix

\subsection{Additional Related Work}\label{sec:apdx-rw}

In this section find additional related work and discussion or previously mentioned related work.

Additional approaches for interpreting neural networks include Feature Leveling~\cite{lu2019not} and Saliency Graphs ~\cite{harel2007graph,nikulin2019free}, involve varying the weights of the network and drawing conclusions about the meaningful features. 
Formal verification can be performed when the state space of the algorithm (the space of all possible inputs to the network) can be fully enumerated~\cite{kazak2019verifying}, or by using other verification techniques that interrogate a black box, such as inductive synthesis~\cite{zhu2019inductive}.
One of the goals of interpretation is predictability and safety.  This can be approached in a different way, by incorporating constraints into the learning procedure, as in Constrained Policy Optimization~\cite{achiam2017constrained}.  

Autonomous systems must be able to deal with uncertain and unforeseen scenarios (risk reducations). One attempt to incorporate risk into RL is to include a risk penalty in the reward function~\cite{kamran2020risk}. Combining planning and RL (with a planner able to override, influence, or augment a learned policy) is a way to incorporate existing risk-based methods into learning methods, affecting executed actions~\cite{mokhtari2021don,hoel2020tactical,angelov2019dynoplan,jin2019risk}. Similarly, safety constraints can be learned and used to affect action selection at execution time~\cite{zhang2020cautious}. Safety constraints can also be used at training time (restricting or influencing action choices during training based on calculated risk) in order to train in the real world with less risk~\cite{kahn2017uncertainty}. In contrast to the above methods, we combine autonomous learning with human domain knowledge.  \cite{liang2021vo} models the Gaussian distributions of the errors in sensor measurements and object detection, but this modeling might be very conservative when the variances are high. \cite{nyberg2021risk, gleirscher2019risk} estimate the severity of safety violations regarding state estimations. Risk management models are intended to analyze other models and systems and identify which subcomponents or sub-models are safe or dangerous.
This risk measurement should also be also computationally efficient \cite{nyberg2021risk} to give the real-time evaluations.

Safety-critical applications can make use of decision trees, because (in addition to the benefits discussed in Section \ref{sec:rw}, numerous existing methods~\cite{de2008z3} can analyze decision trees from a safety and verification standpoint~\cite{blanchet2003static}

One attempt to incorporate risk into RL is to include a risk penalty in the reward function~\cite{kamran2020risk}. Safety constraints can be incorporated by restricting actions during training~\cite{mokhtari2021don,hoel2020tactical,angelov2019dynoplan,jin2019risk} or execution time~\cite{zhang2020cautious}. In contrast to the above methods, we combine autonomous learning with human domain knowledge.

In the planning domain, there is extensive work on risk-aware systems~\cite{primatesta2019risk,hakobyan2019risk,shah2016resolution,primatesta2018risk,jin2019risk}. The same is true when we look at learning methods for autonomous systems, including reinforcement learning.

We choose to extend VIPER ~\cite{VIPER} 
because, unlike other techniques, it is flexible in terms of being able to use any method to learn the expert policy. Moreover, it results in a \textit{single} decision tree, and can provide verifiable guarantees.
First, a neural net policy (or other similar policy) can be learned using reinforcement learning or another method (such as a planning method).  This is considered the expert policy $\pi^*$. (Find a guide to notation in Table \ref{tbl:notation}.) Next, the state-action pairs are generated by executing the policy in the environment. Executing the policy generates trajectories, or sequences of state-action pairs. The state-action pairs from these trajectories are all combined into a large database. From the database  (which includes all the pairs recorded), we in turn sample multiple datasets of state-action pairs. This is a crucial step, upon which our method improves. The Classification and Regression Tree (CART) method~\cite{lewis2000introduction} creates a decision tree policy $\hat \pi_i$ from these datasets. In this supervised learning procedure, the elements of the state in each pair are the features of the tree and the action is the label. For each tree policy created, multiple trials are conducted and the reward is scored for each. The decision tree policy with the highest average reward is considered the best policy $\hat \pi$, which should best duplicate the expert policy. The structure of $\hat \pi$ is that of a binary tree. Each branch node evaluates a conditional expression regarding the feature space (which corresponds to the state space). Each leaf node class corresponds to one of the discrete actions.

Similar approaches have also been used to analyze Networking Systems~\cite{meng2020interpreting} and Carla simulations~\cite{chen2020learning}. 
MoET ~\cite{vasic2019mo} is another work that seeks to improve on VIPER by using a mixture of multiple trees instead of a single tree. Our approach is motivated by these developments and is designed to handle complex tasks corresponding to systems with high degrees-of-freedom or challenging environments (e.g., a robot navigating through a dense crowd) using a single-tree approach.

\subsection{Further Notes on Environments}\label{sec:apdx-env}

In all cases, the state space includes the goal location in polar coordinates (from the perspective of the robot), the previous action taken, and the physical space around the robot as sensed by the lidar (and preprocessed into an occupancy grid). The lidar we use scans $512$ ranges from $-\frac{2}{3}\pi$ to $\frac{2}{3}\pi$ radians (with $0$ corresponding to straight ahead). In our preprocessing step this gets reduced to a radial occupancy grid. The dimensions of the occupancy grid can vary, and others could use different sizes than we chose. In our implementation we used 10 columns of equal radians and 7 rows, beginning 10 cm from the center-point of the robot and with heights of (listed in order from closest to most distant from the robot) $0.2$ m, $0.2$ m, $0.2$ m, $0.3$ m, $1$ m, $1$ m, $1$ m.  The occupancy grid information from the current timestep and the prior two timesteps is included in the state space.
Therefore there are 210 features indicating obstacle position and movement, 2 features describing relative goal position, and 1 feature noting the action chosen by the agent in the previous timestep (total of 213 dimensions in the state space). We use a small discrete action space with motion primitives with varying discrete forward or angular velocities.

We have a simulated and real-world setup for the Indoor Obstacle Avoidance and Outdoor Navigation environments, and simulated environment for Indoor Find \& Follow.

\subsection{Tree Modification Algorithms}\label{sec:apdx-treemod}

Unlike some other approaches that incorporate human input (such as ~\cite{ning2021survey}), in our approach we use human input to correct the errors while keeping the robot a fully autonomous entity at execution time. In this manner, we enable a synthesis of human and machine intelligence.



\subsubsection{Freezing}
A common issue in robot learning for navigation is the ``Freezing Robot Problem.''~\cite{trautman2010unfreezing,trautman2015robot,sathyamoorthy2020frozone,fan2019getting}
Here the robot is presented with a set of obstacles and chooses to remain immobile. The purpose of choosing to remain motionless is to avoid crashing into an obstacle. However, it is also clearly undesirable since the robot has ceased to progress in the direction of the goal. One critical risk is that in the case of static obstacles causing this error, the failure case is one from which the robot cannot extricate itself.

The following procedure identifies nodes in a decision tree that cause (or could potentially cause) the freezing error, and then a subsequent procedure changes the attributes of the identified nodes to reduce the risk of the freezing error. The algorithm for determining which nodes are at fault is found in Algorithm \ref{alg:detect_freezing}

\RestyleAlgo{ruled}
\LinesNumbered
\begin{algorithm}
\textbf{Detect Freezing Nodes}( $\pi^\dag,  a_F, m_A$ ):\\
$\mathcal{N} \gets \varnothing$\;
\For {node $n \in \pi^\dag$}{
    \If {n\text{ is a leaf node}}{
        $m_C \gets$ the number of cells in the occupancy grid in which movement occurs\;
        \If {$m_C < m_A$ and $n[\text{action}] = a_F$} {
            Add $n$ to $\mathcal{N}$\;
        } 
    }
}
Return $\mathcal{N}$
 \caption{ Detect Freezing \label{alg:detect_freezing}}
\end{algorithm}

where $\pi^\dag$ is the tree policy, $a_F$ represents an action or collection of actions corresponding to a ``stop'' action, and $m_a$ is a configurable parameter indicating ``the maximum number of polar occupancy grid cells that can contain movement while still considering the obstacles as static.'' (See Appendix Section \ref{sec:apdx-env} for an explaination of the occupancy grid.)
The algorithm analyzes each leaf node in the tree. If the node's action is the Stop action, and if the occupancy grid described by the state bounds of that node indicate stationary obstacles (within a tolerance controlled by $m_A$), then the node is considered to be a problematic node and is added to the list of potential-freezing nodes. (Sometimes we may not want to be too strict about disallowing movement among the obstacles, necessitating the $m_A$ parameter. Setting $m_A = 0$ means that a node will not be considered problematic unless the obstacles are perfectly still, and setting $m_A$ at the maximum will cause the algorithm to place all nodes with the stop action $a_F$ into the problematic set regardless of the obstacle movement and position implied by the state subspace. By ``subspace'' we mean a bounded subset of the state space.) 
An edge case is when the node's subspace dimensions encompass both moving and non-moving possibilities. (In other words, the boundaries of a leaf node include states where a particular occupancy grid cell is static and states where that particular occupancy grid cell indicates movement over the three timesteps.) In this case, the algorithm will evaluate the condition as true (``stationary'') so long as the bounds of those dimensions remain the same for all timesteps. (This is because even though movement might occur sometimes, the situation where an obstacle is in fact motionless is included in this subspace. In other words, while there might be some movement in a particular cell associated with this leaf node, a static obstacle in that cell is also associated with this same leaf node, and so a problematic state-action pair is associated with this leaf node even though non-problematic state-action pairs are also associated with it.)
\RestyleAlgo{ruled}
\LinesNumbered
\begin{algorithm}
\textbf{Modify Freezing Nodes}( $\pi^\dag, \mathcal{N}, a_R, a_L$ ):\\
\For {node $n \in \mathcal{N}$}{
    \eIf{majority of obstacles are on the right}{
        $n[\text{action}] \gets a_L$\;
    }{
        $n[\text{action}] \gets a_R$\;
    }
}
Return the updated $\pi^\dag$
 \caption{ Alleviate Freezing \label{alg:fix_freezing}}
\end{algorithm}
The algorithm intended to alleviate this issue is found in Algorithm \ref{alg:fix_freezing},
where $a_R$ and $a_L$ are actions corresponding to pure right and left rotation (no linear velocity), respectively. This should safely allow the robot to find an observed state where it can extract itself from stasis. (This may result in a non-optimal path, but it will not be less optimal than failing to complete the task completely. Some paths that do complete the task may be made slightly less optimal as a result of these changes. In the real world, there are times when this tradeoff will be desired.)

\subsubsection{Oscillation}

\RestyleAlgo{ruled}
\LinesNumbered
\begin{algorithm}
\textbf{Detect Oscillation Nodes}( $\pi^\dag, E, \mathcal{F}_O, n_e, L$ ):\\
$\mathds{H}^L \gets $ initialize an empty queue\;
$\mathcal{N} \gets \varnothing$\;
$\mathcal{O}_C(i) \gets \varnothing \quad \forall \quad i \in \{$ ids of nodes in $\pi^\dag\}$\;
$\mathcal{O}_X(i) \gets \varnothing \quad \forall \quad i \in \{$ ids of nodes in $\pi^\dag\}$\;
\For{$n_e$ episodes}{
    Reset environment $E$\;
    \While{E does not indicate episode done}{
        $s \gets$ get current state from $E$\;
        $a \gets \pi^\dag(s)$\;
        Append $(s, a)$ to $\mathds{H}^L$, removing the oldest if the length of the queue is $> L$\;
        \eIf{$\mathcal{F}_O(\mathds{H}^L)$}{
            \{$n_i, n_{i+1}, ..., n_{i+L}\} \gets $ leaf nodes in $\pi^\dag$ corresponding to each $s \in \mathds{H}^L$\;
            Add $\{n_i, n_{i+1}, ..., n_{i+L}\}$ to $\mathcal{N}$\;
            Add all $s \in \mathds{H}^L$ to $\mathcal{O}_C(d_i), \mathcal{O}_C(d_{i+1}), ..., \mathcal{O}_C(d_{i+L})$, where $d_d$ is the corresponding id of each node in $\{n_i, n_{i+1}, ..., n_{i+L}\}$\;
        }{
            $n_i \gets \text{ leaf node in $\pi^\dag$ corresponding to } s$\;
            Add $s$ to $\mathcal{O}_X(d)$, where $d$ is the id of $n_i$\;
        }
        Execute action $a$ in environment $E$\;
    }
}
Return $\mathcal{N}, \mathcal{O}_C(i), \mathcal{O}_X(i)$\;
 \caption{ Detect Oscillation \label{alg:detect_oscillation}}
\end{algorithm}
The next issue we address with regards to correcting the imperfect expert policies is the issue of oscillation.  While traveling around obstacles, the robot will occasionally alternately rotate too far towards and away from the impediment. This behavior is inefficient and off-putting to a human observer. As before, we develop a procedure to mitigate the issue. We run the policy in simulation, records the robot's behavior, and note which parts of the decision-tree policy are activated when the undesirable behavior occurs (contributing to the behavior).  We modify the tree by modifying existing nodes or adding nodes, replacing existing actions with replacement actions that have lower linear and angular velocities. The procedure for detecting the nodes responsible for the undesirable behavior is found in Algorithm \ref{alg:detect_oscillation},
where $E$ is an environment, $\mathcal{F}_O$ is a function that accepts a state-action pair sequence and returns a boolean describing whether the trajectory sequence experienced oscillation, $L$ is the sequence length, and $n_e$ is the total number of episodes to monitor.
The procedure to mitigate the excessive oscillation is found in Algorithm \ref{alg:fix_oscillation}. 
\RestyleAlgo{ruled}
\LinesNumbered
\begin{algorithm}
\textbf{Alleviate Oscillation Nodes}( $\pi^\dag, \mathcal{N}, \mathcal{O}_C, \mathcal{O}_X, z$ ):\\
// Note that all $n \in \mathcal{N}$ are in $\pi^\dag$\\
\For{$n_i \in \mathcal{N}$}{
    \eIf{$\mathcal{O}_X(i) = \varnothing \text { or } z$}{
        // All states visited on this node are oscillation\\
        $n_i$[action] $\gets$ action with linear and angular velocity of reduced magnitude\;
    }{
        $X \gets$ new leaf node with action $n_i$[action]\;
        $C \gets$ new leaf node with action with linear and angular velocity of reduced magnitude compared to $n_i$[action]\;
        $n_i$ is turned into a branch node with $X$ and $C$ as children, splitting on the best split that best separates the states in $\mathcal{O}_C(i)$ to node $C$ and states in $\mathcal{O}_S(i)$ to node $X$\;
    }
}
Return updated $\pi^\dag$\;
 \caption{ Alleviate Oscillation \label{alg:fix_oscillation}}
\end{algorithm}
Each node that has a subspace corresponding to an observation of oscillation is given two children nodes (and itself becomes a branching node). One of the children corresponds to the problematic subspace and the other corresponds to the subspace where no problems were observed. The latter node is given the action of the original node. The new child node corresponding to the problematic behavior sub-space is assigned a lower magnitude velocity action.
In Algorithm \ref{alg:fix_oscillation}, each $\mathcal{O}_C(i) \in \mathcal{O}_C$ is a set of states in state subspace of node $i$ where oscillation occurs, each $\mathcal{O}_X(i) \in \mathcal{O}_X$ is a set of states in state subspace of node $i$ where oscillation does not occur, and $z$ is a boolean.

\begin{table*}
    \centering
    {\resizebox{1.0\textwidth}{!}{
    \begin{tabular}{|l|c|c|c|c|c|c|c|c|c|c|c|c|c|c|c|}
        \hline
         \textbf{Action ID} & 0 & 1 & 2 & 3 & 4 & 5 & 6 & 7 & 8 & 9 & 10 & 11 & 12 & 13 & 14 \\
         \hline
         \textbf{Linear Velocity} & 1 & 0 & 1 & 0 & 1 & 0 & 0.4 & 0 & 0 & 1 & 1 & 0.4 & 0.4 & 0.4 & 0.4 \\
         \textbf{Angular Velocity} & -1 & -1 & 0 & 0 & 1 & 1 & 0 & -0.4 & 0.4 & -0.4 & 0.4 & -1 & 1 & -0.4 & 0.4 \\
         \hline
    \end{tabular}
    }
    \caption{Expanded Action Space 2, used in the outdoor scenarios only}}
    \label{tble:expanded_action space}
\end{table*}

\subsubsection{Vibration}

We introduce two tree modifications aimed at addressing excessive vibration.  These can be used to address vibration when the learned policy does not sufficiently reduce vibration below the threshold, or when one wants to modify the vibration threshold after the fact without retraining.

\RestyleAlgo{ruled}
\LinesNumbered
\begin{algorithm}
\textbf{Detect Vibration Relevant Nodes}( $\pi^\dag,  \mathcal{V}_E$ ):\\
$\mathcal{N} \gets \varnothing$\;
\For {node $n \in \pi^\dag$}{
    \If {n\text{ is not a leaf node}}{
        \If {$n[\text{feature}] \in \mathcal{V}_E$} {
            Add $n$ to $\mathcal{N}$\;
        } 
    }
}
Return $\mathcal{N}$
 \caption{ Detect Vibration Method 1 \label{alg:detect_vib1}}
\end{algorithm}

The first approach involves treating the threshold of nodes that split on a feature corresponding to angular velocity as a hyperparemeter, and modifying it to produce the desired result.

To identify relevant nodes, see the procedure in Algorithm \ref{alg:detect_vib1}.  In our case, $\mathcal{V}_E = \{872,873, 883, 884, 894, 895, 905, 906\}$, and represents the feature indices of features related to angular velocity ($\omega_r$ and $\omega_p$ at all timesteps present in the state).  Any node in the tree that splits on one of these features is a relevant node.

In Algorithm \ref{alg:fix_vib1}, we show the procedure for adjusting the threshold at each of these nodes by an increment $h$.  As noted, $h$ is a hyperparameter that the human user can try out (since the tree modification is very fast) and tune to the desired level, measuring the vibration after each change.

\RestyleAlgo{ruled}
\LinesNumbered
\begin{algorithm}
\textbf{Modify Vibration-Relevant Nodes}( $\pi^\dag, \mathcal{N}, h$ ):\\
\For {node $n \in \mathcal{N}$}{
    $n[\text{threshold}] \gets n[\text{threshold}] + h$\;
}
Return the updated $\pi^\dag$
 \caption{ Reduce Vibration Method 1 \label{alg:fix_vib1}}
\end{algorithm}

There is a second Vibration reduction method that looks at the equation
\begin{equation}\label{eqn:Vb2}
    V_b = \sum_{t'=t-3}^t \gamma^{t-t'} ( |{\omega_r}|+| \omega_p|)
\end{equation}
and notes that this also describes a closed hyperdimensional object.  Specifically, for the four timesteps and two angular velocity values present above, it describes an eight-dimensional object. This object is a hyperrectangle and a convex hull.  We call this the "Vibration Space." 


\RestyleAlgo{ruled}
\LinesNumbered
\begin{algorithm}
\textbf{Detect Vibration-Space-Intersecting Nodes}( $\pi^\dag,  \mathcal{V}_E, V_b, \gamma$ ):\\
$\mathcal{N}_V \gets \varnothing$\;
Let $\mathcal{H}_V$ be a set of points in a hyperspace with dimensions equal to the size of $\mathcal{V}_E$. These points form a convex hull describing the space taken by Equation \ref{eqn:Vb2}, which we call the Vibration Space\;
$\mathcal{C}_N \gets$ a set of nodes initially containing only the root node\;
\While{$\mathcal{C}_N \neq \varnothing$}{
    $n \gets$ remove the last node from $\mathcal{C}_N$\;
    Let $\mathcal{H}_N$ be a set of points forming a convex hull that is the boundaries of the abstract state space of $n$, considering only the dimensions indicated by $\mathcal{V}_E$\;
    \eIf{surfaces of $\mathcal{H}_N$ and $\mathcal{H}_V$ intersect}{
        Add $n$ to $\mathcal{N}_V$\;
    }{
        Add both children of $n$, if any, to $\mathcal{C}_N$\;
    }
}
Return $\mathcal{N}_V$
 \caption{ Detect Vibration Method 2 \label{alg:detect_vib2}}
\end{algorithm}

Consider Algorithm \ref{alg:detect_vib2}. The first two arguments to the procedure are the same as in Algorithm \ref{alg:detect_vib1}, and $V_b, \gamma$ come from the vibration space equation. In a tree, the root node's abstract state space encompasses the entire state space. It's two children each encompass a mutually exclusive subset of the state space, and so on. We consider the relationship between the subset of the state space corresponding to a node in the tree and the vibration space. The vibration space might be outside of the state space, fully contained by the state space, or intersect the boundary of the state space.  We posit that if the vibration space intersects the boundary of the node state space, we can use that as a proxy for the policy making a decision based on that vibration. Potentially, the decision it makes not enough to dampen the vibration. In Algorithm \ref{alg:detect_vib1}, we want to find all nodes where the surface of the vibration space intersects with the surface of the node's state subspace.  (Note we want to check for intersection of surface of the shapes and not intersection of the volumes of the shape because there will always be at least one node at every depth with contains the vibration space within it's state subspace, even if the vibration space is nowhere near the boundary.

\RestyleAlgo{ruled}
\LinesNumbered
\begin{algorithm}
\textbf{Modify Children of Vibration-Relevant Nodes}( $\pi^\dag, \mathcal{N}_V, \mathbf{M}_c$ ):\\
\For {node $n_V \in \mathcal{N}_V$}{
    \eIf{$n_V$ is a leaf node}{
        Let $\mathcal{N}_L$ be a set of nodes containing $n_V$ only\;
    }{
        Let $\mathcal{N}_L$ be a set of nodes containing all leaf node descendants of $n_V$\;
    }
    \For {node $n_L \in \mathcal{N}_L$}{
        $n[\text{action}] \gets \mathbf{M}_c[n[\text{action}]]$\;
    }
}
Return the updated $\pi^\dag$
 \caption{ Reduce Vibration Method 2 \label{alg:fix_vib2}}
\end{algorithm}

In Algorithm \ref{alg:fix_vib2}, see the procedure for modifying the tree based on the identified nodes. The algorithm makes a modification to the action of all of the leaf nodes descended from the nodes which have boundaries intersecting the vibration space.  In general, to reduce vibration, magnitude of velocities is decreased.  The specific mapping is provided by $\mathbf{M}_c : T \rightarrow T$, where $T$ is the action type (in our case integer).  In our case, we used $\mathbf{M}_c = \{ 0 \rightarrow 13; 1 \rightarrow 7; 2\rightarrow 6; 3\rightarrow 3; 4\rightarrow 14; 8\rightarrow 6; 7\rightarrow 8; 9 \rightarrow 13; 10 \rightarrow 14; 11 \rightarrow 13; 12 \rightarrow 14; 13 \rightarrow 13; 14 \rightarrow 14 \} $, with the action space described in Table V.



\subsection{Limitations}\label{sec:apdx-limitations}


MSVIPER requires a discrete action space and a state space in the format of a numerical array. 
In addition, MSVIPER (like its predecessors) uses hyperparameters that must be tuned for optimal results. These include sample size $n_S$, number of iterations $n$, and trajectories generated per iteration $m$. There may be tasks that a neural net can solve but which requires such a large tree that it is computationally infeasible, such tasks would not be a good fit for MSVIPER or related methods. Finally, since MSVIPER's first step is an imitation process, there is a chance that it needs to deal with the issues  of distribution shifting. 

In our indoor navigation environment with obstacles, we limited obstacles to be static or dynamic with relatively slow speed, which could be captured by robots' sensors. If the robot cannot sense the obstacle quickly enough, it will not be able to avoid it. One of the benefits of the tree modification is that it allows for incorporating human knowledge. However, it does require a human to encode that knowledge explicitly in a program in order to capture it.

\subsection{MSVIPER Analysis: Full Proofs}\label{sec:app-full-proofs}


In the following sections, we analyze the upper bound on divergence between the tree and the expert policy in the cases of MSVIPER and VIPER, and we compare expectations for sample complexity of VIPER and MSVIPER.

\subsubsection{Upper Bound on Divergence Between Expert and Learned Policy}

Our work makes the decision tree efficiently and effectively mimic the expert policy. If the loss function between the expert policy and the decision tree is smaller, the decision tree imitates the expert policy better. Therefore if the upper bound of the loss function is smaller, the decision tree would be guaranteed to have a better imitating performance. 

\begin{theorem}
For the multi-stage MSVIPER, the loss function is :
\begin{equation}
        \Tilde{l}_{t}(s,\pi) = V_{t}^{(\pi ^*)}(s)-\min_{a \in A}\hat Q_{t}^{(\pi ^*)}(s,a).
\end{equation}
where
\begin{equation}\label{th:final1}
    \hat Q^{\pi^*}(s,a) = 
    \mathbb{E}_{\pi}
    \left[\sum_{e \in E}\left(\sum_{k=0}^{\infty}\gamma^{k}R^e_{t+k+1}|S_{t}=s, A_{t}=a\right) \cdot w_e\right]
\end{equation}
\label{theorem:loss_function}
\begin{proof}
    According to \cite{VIPER}, the loss function is:
    \begin{equation}
        l(\pi) = T^{-1}\sum_{t=0}^{T-1}\mathbb{E}_{s \sim d_{t}^{(\pi)}}[l_{t}(s,\pi)]
    \end{equation}
    where T is the time horizon (maximum timesteps) and the loss $l_{t}(s,\pi)$ is:
     \begin{equation}
        l_{t}(s,\pi)=V^{\pi^{*}}_{t}(s)-Q^{\pi^{*}}_{t}(s,\pi(s)).
    \end{equation}
    where $V(s)$ is the value function of the policy and $Q(s,a)$ is the quality function. 
    In order to guarantee a better performance, we need to decrease the upper bound of $\Tilde{l}_{t}(s,\pi)$.  
    After training by training set $E$ we have oracle policy $\pi^*$. According to the Q-Dagger algorithm, the hinge loss function $l(\pi)$ is upper bounded by a convex loss $\Tilde{l}_{t}(s,\pi)$:
    \begin{equation}
        \Tilde{l}_{t}(s,\pi) = V_{t}^{(\pi ^*)}(s)-\min_{a \in A}Q_{t}^{(\pi ^*)}(s,a).
    \end{equation}
    When we train the decision tree in multiple environments $e \in E$, considering each environment has a different contribution to policy, we have the weight for each environment $e$:
    \begin{align}
        \omega_e = f(R^e,T^e)
    \end{align}
    Here $\sum_{e\in E}\omega_e=1$, $R_i^e$ is the reward at timestep $i$ for environment $e$, and $T^e$ is the total timesteps of trajectories in each scenario $e$. Then the quality function $\hat Q^{\pi^*}(s,a)$ of multiple stages could be:
    \begin{align}
        \hat Q^{\pi^*}(s,a) = \sum_{e \in E} Q^{\pi^*}(s,a) \cdot \omega_e
    \end{align}
    \begin{equation}\label{th:final1}
        \hat Q^{\pi^*}(s,a) =  \mathbb{E}_{\pi}
        \left[\sum_{e \in E}\left(\sum_{k=0}^{\infty}\gamma^{k}R^e_{t+k+1}|S_{t}=s, A_{t}=a\right) \cdot w_e\right]
    \end{equation}
    Replace $Q^{\pi^*}(s,a)$ with $\hat Q^{\pi^*}(s,a)$ we have:
    \begin{equation}
        \Tilde{l}_{t}(s,\pi) = V_{t}^{(\pi ^*)}(s)-\min_{a \in A}\hat Q_{t}^{(\pi ^*)}(s,a).
\end{equation}
\end{proof}
\end{theorem}

According to \cite{VIPER}, while in training, the loss function can be calculated by:
\begin{equation}
    Tl(\pi) = J(\pi) - J(\pi^*)
\end{equation}
where $\pi$ is any policy trained by the decision tree, and $\pi^*$ is the policy optimally imitating expert policy. For multiple scenarios, we obtain:
\newtheorem{lemm2}{Lemma}
\begin{lemm2}\label{lemma:l1}
    Let $\pi^{*}$ be held constant, Equation \eqref{eq:loss_boundance} holds when using samples from different scenarios.
    \begin{equation}\label{eq:loss_boundance}
    J(\hat{\pi})\le J(\pi^{*})+T\epsilon_{N}+\Tilde{O}(1)
    \end{equation}
\end{lemm2}
\begin{proof}
    \cite{VIPER} proposed equation \eqref{eq:loss_boundance} for single stage training. (See ~\cite{VIPER,ross2011reduction} for more details on the $\Tilde{O}(1)$.) According to \cite{ross2011reduction}, for each trajectory in different scenarios $J(\pi)=T \mathbb{E}_{s\sim d^{t}_{\pi}}[C_{\pi}(s)]$ holds. Since sampled trajectories are independent, we can simply add the trajectories of different scenarios into states distribution class D, then the function $\epsilon_{N}$ holds.
    As a result, the policy $\hat\pi$ satisfies equation \ref{eq:loss_boundance}.
\end{proof}
As per Lemma \ref{lemma:l1}, the difference between an expert policy and a well-trained policy by MSVIPER is also upper bounded. In the next section we will prove that MSVIPER can decrease the upper bound of a loss function, and thus that MSVIPER has a better performance than VIPER.

\subsubsection{Sampling Complexity of MSVIPER}

\begin{theorem}
$P_M(\epsilon|S_C, l, m, E) \geq  P_V(\epsilon|S_C, l, m)$ and $u_M \leq u_V$ (MSVIPER has a sample complexity equivalent or superior to that of VIPER) if $\overline{\mathbb{S}}_{k}^{e_i,l} < \overline{\mathbb{S}}_{k}^{e_{n_E},l}$ for any $k$ and $i \neq n_E$.
\end{theorem} 

We define ``sampling complexity'' as a measure of how many state-action pair tuples (and thus trajectories generated) are required from the expert policy to learn a decision tree that achieves a given threshold of performance (as measured by average reward achieved compared to the expert policy).

We assert that, given certain assumptions stated in this section, MSVIPER has a sample complexity as good as or better than VIPER.  This is shown when the upper bound $u$ for the divergence in reward between the expert policy and the tree policy is, in the MSVIPER case, smaller than or equivalent to the VIPER case.

Sample complexity is improved when the likelihood that high accuracy is achieved on critical states increases.


We look at this upper bound $u$, described above and in ~\cite{VIPER} as 
\begin{equation}
\begin{split}
    Q_t^{(\pi^*)}&(s, a) - Q_t^{(\pi^*)}(s, \pi^*(s)), \\
    &\text{ for all } a \in A, s \in S, \text{ and } t \in {0, ..., T-1}.
\end{split}
\end{equation}


It is noted that $u$ ``may be $O(T)$ ...if there are \textit{\textbf{critical states}} such that failing to take action $\pi^*(s)$ in state $s$ results in forfeiting all subsequent rewards.'' This is the case, for example, for the navigation environment, where a collision results in an immediate failure. The states where possible collision is imminent are the ``critical states.'' The same principal can be applied to other domains.  In this instance $u = O(T)$.  It is further noted that this bound is $O(T)$ ``as long as $\hat{\pi}$ achieves high accuracy on critical states.''   It follows that any improvement to the method that results in increasing the likelihood that high accuracy is achieved on critical states improves the likelihood of the bound being $O(T)$ as a opposed to a higher value.

Let $S_C \in S$ be the set of all critical states. Let $k_i \in S_C$ for $i = 1 \text{ to } K$ 
be a list of all critical states.  We assume that the definition of whether a state is considered a critical state is the same across scenarios.  We assume that $S_C$ is a finite set.

Consider the sampling step of our algorithm

\begin{equation}
  D_i \gets \{ (s_j, \pi^*(s_j)) \sim d^{s_{j-1}} \}  
\end{equation}

Assume that the expert policy $\pi^*$ 
chooses correct actions 
for critical states, and that the decision tree $\hat{\pi}$ could thus only learn to fail on critical states if it fails to sample from critical states during trajectory generation. 

Let $\mathbb{S}_k^{e,l}$ be a binary random variable, with the associated probability distribution being the Bernoulli distribution with probability $p = p_k^{e, l}$ (as in, $P(\mathbb{S}_k^{e,l}) = p_k^{e, l}$). (Note $0 \leq p_k^{e, l} \leq 1$). For a given generation of a trajectory of length $l$ drawn from  environment $e$, ($\mathbb{S}_k^{e,l}$ == 1) represents the outcome that the critical state $k$ is sampled as part of the trajectory.  ($\mathbb{S}_k^{e,l}$ == 0) represents the outcome that the critical state $k$ is not among the trajectory (although other critical states may be).  Let $\overline{\mathbb{S}}_k^{e,l}$ be a similar binary random variable, where an outcome of $({\overline{\mathbb{S}}}_k^{e,l} ==1)$ means that state $k$ is not among the states within the drawn trajectory.



Let $P(\epsilon|S_C, l)$ be the probability that the number of critical states present in a drawn trajectory of length $l$ is a fraction of the total number of critical states $S_C$ equal to or greater than $\epsilon$, for $0 < \epsilon < 1$.


The number of critical states drawn from a trajectory of length $l$ from environment $e$ can be represented by the sum $\mathbb{S}_{k_1}^{e,l} + \mathbb{S}_{k_2}^{e,l} + ... + \mathbb{S}_{k_K}^{e,l} $ and the fraction of total states drawn is 
\begin{equation}
 \frac{\mathbb{S}_{k_1}^{e,l} + \mathbb{S}_{k_2}^{e,l} + ... + \mathbb{S}_{k_K}^{e,l} }{K}   
\end{equation}

For a single environment $e$, 
\begin{equation}
    P(\epsilon|S_C, l) = 
    P\left( \frac{  \mathbb{S}_{k_1}^{e,l} + \mathbb{S}_{k_2}^{e,l} + ... + \mathbb{S}_{k_K}^{e,l} }{K} \geq \epsilon \right)
\end{equation}

For single-scenario VIPER using a training that ends up incorporating $m$ total trajectories, the probability of a fraction of the total critical states being $\geq \epsilon$ is

\begin{equation}
\begin{split}
    P_V&(\epsilon|S_C, l, m)  =  \\
    &=P \left( \frac{1}{K} \sum_{k \in S_C} \left( \mathbb{S}_{k}^{e,l} \right) \geq \epsilon \right) \text{ repeated $m$ times } \\
    &=P \left( \frac{1}{K} \sum_{k \in S_C} \left( \sum_{j=1}^{m} X_j \sim \mathbf{B}\left(m, p_k^{e, l}\right) \right) \geq \epsilon \right) \\
\end{split}
\end{equation}

where $\mathbf{B}(m, p)$ is the Binomial distribution.

For MSVIPER using a training that ends up incorporating $m$ total trajectories evenly divided across environments $e_1, ... e_{n_E} \in E$ (assuming $m$ must be evenly divisible by $E$),  the probability of a fraction of the total critical states being $\geq \epsilon$ is

\begin{equation}
\begin{split}
    P_M&(\epsilon|S_C, l, m, E)  =  \\
    &=P \left( \frac{1}{K} \sum_{k \in S_C} \left(1 -    \prod_{i=1}^{n_E}\left(
    \overline{\mathbb{S}}_{k}^{e_i,l} \right) \right)  \geq \epsilon \right) \text{ repeated $m$ times } \\
    &=P \left( \frac{1}{K} \sum_{k \in S_C} \left( \sum_{j=1}^{m} X_j \sim \mathbb{B} \right) \geq \epsilon \right), \\
    & \qquad \qquad \mathbb{B} =  \mathbf{B}\left(\frac{m}{n_E}, 1- \left(p_k^{e_1, l } \cdot p_k^{e_2, l } \cdot ... \cdot p_k^{e_{n_E}, l }  \right) \right)\\
\end{split}
\end{equation}

Note that $P_V(\epsilon|S_C, l, m) =  P_M(\epsilon|S_C, l, m, E)$ if $E$ consists only of $e_{n_E}$.

If any $\overline{\mathbb{S}}_{k}^{e_i,l} < \overline{\mathbb{S}}_{k}^{e_{n_E},l}$ for any $k$, for any $i \neq n_E$ (where $e_{n_E}$ indicates the final stage) then we can say that 

\begin{equation}
     P_M(\epsilon|S_C, l, m, E) \geq  P_V(\epsilon|S_C, l, m) 
\end{equation}

Given this relationship between the probabilities, we can say that for any $\epsilon$ where the assumptions noted above hold, the probability that MSVIPER utilizes this $\epsilon$-fraction of critical states in its training of the tree is greater than or equal to the probability that VIPER does so.  Thus, under these conditions the upper bounds $u$ for the divergence between expert and tree are equivalent between VIPER and MSVIPER, or MSVIPER has a smaller bound. ($u_M \leq u_V$)  Thus, MSVIPER has equivalent or superior sample complexity, provided that  any $\overline{\mathbb{S}}_{k}^{e_i,l} < \overline{\mathbb{S}}_{k}^{e_{n_E},l}$ for any $k$ and $i \neq n_E$.




\end{document}